\documentclass{sn-jnl}

\jyear{2023}%

\theoremstyle{thmstyleone}%
\theoremstyle{thmstyletwo}%

\theoremstyle{thmstylethree}%
\usepackage{comment}
\usepackage{graphicx}
\usepackage[thinlines]{easytable}
\usepackage{tabulary}
\usepackage{array}

\usepackage{array}

\newcolumntype{L}[1]{>{\raggedright\let\newline\\\arraybackslash\hspace{0pt}}m{#1}}
\newcolumntype{C}[1]{>{\centering\let\newline\\\arraybackslash\hspace{0pt}}m{#1}}
\newcolumntype{R}[1]{>{\raggedleft\let\newline\\\arraybackslash\hspace{0pt}}m{#1}}

\newcolumntype{L}[1]{>{\raggedright\let\newline\\\arraybackslash\hspace{0pt}}m{#1}}
\newcolumntype{C}[1]{>{\centering\let\newline\\\arraybackslash\hspace{0pt}}m{#1}}
\newcolumntype{R}[1]{>{\raggedleft\let\newline\\\arraybackslash\hspace{0pt}}m{#1}}
\usepackage{ragged2e}
\newcolumntype{P}[1]{>{\RaggedRight\hspace{0pt}}p{#1}}
\usepackage{longtable}
\usepackage{lipsum}
\usepackage[utf8]{inputenc}

\begin{document}

\title[ ]{Crime Prediction Using Machine Learning and Deep Learning: A Systematic Review and Future Directions}


\author*[1]{\fnm{Varun} \sur{Mandalapu}}\email{varunm1@umbc.edu}
\equalcont{These authors contributed equally to this work.}

\author[2]{\fnm{Lavanya} \sur{Elluri}}\email{elluri@tamuct.edu}
\equalcont{These authors contributed equally to this work.}

\author[2]{\fnm{Piyush} \sur{Vyas}}\email{piyush.vyas@tamuct.edu}
\equalcont{These authors contributed equally to this work.}

\author[1]{\fnm{Nirmalya} \sur{Roy}}\email{nroy@umbc.edu}

\affil*[1]{\orgdiv{Information Systems}, \orgname{University of Maryland Baltimore County}, \orgaddress{\street{1000 Hilltop Cir}, \city{Baltimore}, \postcode{21250}, \state{MD}, \country{USA}}}

\affil[2]{\orgdiv{Computer Information Systems}, \orgname{Texas A\&M University - Central Texas}, \orgaddress{\street{1001 Leadership Pl}, \city{Killeen}, \postcode{76549}, \state{TX}, \country{USA}}}


\abstract{Predicting crime using machine learning and deep learning techniques has gained considerable attention from researchers in recent years, focusing on identifying patterns and trends in crime occurrences. This review paper examines over 150 articles to explore the various machine learning and deep learning algorithms applied to predict crime. The study provides access to the datasets used for crime prediction by researchers and analyzes prominent approaches applied in machine learning and deep learning algorithms to predict crime, offering insights into different trends and factors related to criminal activities. Additionally, the paper highlights potential gaps and future directions that can enhance the accuracy of crime prediction. Finally, the comprehensive overview of research discussed in this paper  on crime prediction using machine learning and deep learning approaches serves as a valuable reference for researchers in this field. By gaining a deeper understanding of crime prediction techniques, law enforcement agencies can develop strategies to prevent and respond to criminal activities more effectively. }




\maketitle

\section{Introduction}\label{sec1}

Crime prediction is a complex problem requiring advanced analytical tools to effectively address the gaps in existing detection mechanisms. With the increasing availability of crime data and through the advancement of existing technology, researchers were provided with a unique opportunity to study and research crime detection using machine learning and deep learning methodologies. Based on the recent advances in this field \cite{shah2021crime}\cite{chun2019crime}\cite{kshatri2021empirical}, this article will explore current trends in machine learning and deep learning for crime prediction and discuss how these cutting-edge technologies are being used to detect criminal activities, predict crime patterns, and prevent crime. Our primary goal is to provide a comprehensive overview of recent advancements in this field and contribute to future research efforts.

Machine learning is a subset of artificial intelligence that uses statistical models and algorithms to analyze and make predictions based on data. On the other hand, deep learning is a subset of machine learning that uses artificial neural networks with multiple layers to model complex relationships between inputs and outputs \cite{janiesch2021machine}. Both machine learning and deep learning technologies have the potential to be applied to the problem of crime prediction in many ways.  

Machine learning algorithms have been utilized in crime prediction to analyze crime data and predict future crime patterns. For example, algorithms like decision trees, random forests, and support vector machines have been trained on crime data from specific cities to predict crime patterns accurately \cite{raza2021data}. Apart from predicting crime patterns, these algorithms can provide valuable insights into crime trends and patterns. These capabilities allow for deploying resources and tactics to combat crime effectively. Additionally, machine learning algorithms can also be used to identify correlations between crime incidents and various environmental and demographic factors such as location, weather, and time of day \cite{elluri2019developing}. This information can be used to develop crime prediction and prevention strategies suitable to a given community's specific needs.  

Predictive policing is also a significant application of machine learning for crime prediction \cite{meijer2019predictive}. Predictive policing refers to using data and analytics to inform law enforcement efforts and reduce crime. Machine learning algorithms can be used to analyze crime data from a specific geographic area, such as a city or neighborhood, to identify crime hotspots and predict future crime incidents. This information can then be used to direct policing resources to areas where they are most needed, increasing the effectiveness of law enforcement efforts.  

Deep learning algorithms, such as convolution and recurrent neural networks, have also shown promise in crime prediction. These algorithms have been trained on crime data with either a spatial or temporal component to accurately predict crime patterns in specific cities. For example, deep learning algorithms have been used to analyze crime data, including the time, location, and type of crime incidents \cite{hossain2020crime}. This information is used to create a predictive model that can be used to identify potential crime hotspots and predict future crime incidents.  

Another application of deep learning in crime prediction is computer vision and video analysis. This technology has been used to analyze video footage from surveillance cameras to detect and classify criminal activities, such as vandalism, theft, and assault \cite{shah2021crime}. The advanced deep learning models are also integrated with drones and other aerial technologies to provide new opportunities to monitor and respond to criminal activities. These algorithms have also been used to analyze crime data from multiple sources, including crime reports, social media, and police records, providing a more comprehensive view of criminal activities \cite{saraiva2022crime}. By automating this process, deep learning algorithms have the potential to enhance the ability to identify and respond to crime in real-time, providing a crucial tool in the fight against criminal activity.  

Despite the promise of machine learning and deep learning for crime prediction, several challenges must be addressed. One of the biggest challenges is the availability of high-quality crime data. Crime data can be difficult to obtain, and the available data may need to be completed or reliable. Additionally, collecting and using crime data is associated with privacy and ethical concerns. These challenges must be addressed to fully realize the potential of machine learning and deep learning for crime prediction.  
Another challenge is the interpretability of machine learning and deep learning models. These models can be challenging to understand and interpret, limiting their usefulness in decision-making. To effectively apply these models to the problem of crime prediction, it is vital to develop interpretable models that can provide clear explanations of their predictions.  

Moreover, the recent advancements in machine learning and deep learning for crime prediction show great promise in addressing this complex problem \cite{kounadi2020systematic}. However, significant challenges remain, and much work is still needed to realize these technologies' potential fully. This research article provides a comprehensive overview of recent trends in this field and offers insights into the potential applications of machine learning and deep learning for crime prediction. By highlighting the potential of these technologies and the challenges that must be addressed, this research article contributes to the broader research community. It advances our understanding of the role of machine learning and deep learning in crime prediction. Hence, the key contributions of this work are as follows:- first, this paper provides the amalgamation of existing studies that utilized state-of-the-art machine learning and deep learning-based approaches in the realm of detecting neighborhood crime. Thereby extending the fathomable literature knowledge base. Second, this paper eliminates the limitation of the scarcity of potential datasets availability. We have highlighted distinct publicly available datasets related to neighborhood crime prediction that existing studies have utilized. Thereby archiving the data resources for future scholars. Third, this work drafted future research directions to eliminate the existing research gaps in neighborhood crimes. Thereby reasonably providing future research objectives/questions to the research community to pursue further.

\section{Research Methodology}\label{sec3}

The primary research aims to find various efficient algorithms for predicting neighborhood crimes. In our previous work \cite{elluri2019developing}, we used statistical analysis to predict the crimes in Newyork city. Our paper got good attention from the researchers, so we wanted to look for the efficient machine learning and deep learning approaches used in this area. We have followed a systematic approach to select the papers for this review. As part of this research, we have considered the papers from multiple databases related to predicting crime.   

\begin{figure}[h]%
\centering
\includegraphics[width=0.5\textwidth]{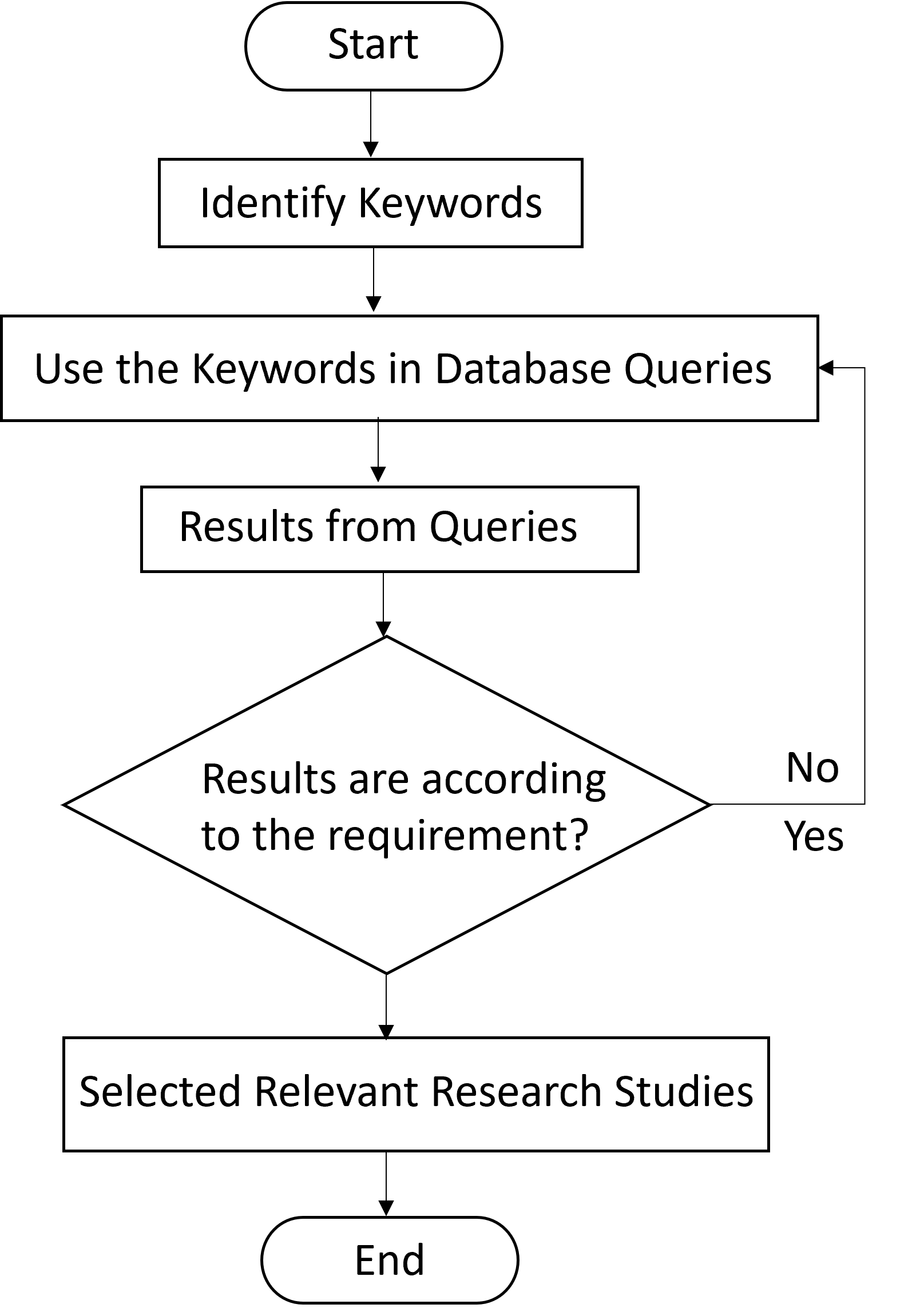}
\caption{Steps involved for typical crime detection}\label{fig1}
\end{figure}

For this review, we have considered all the primarily used terms in the papers focused on predicting crimes. To include all the possible alternative words of each term, we have used “*” as a wild character for IEEE and ACM databases so that it contains zero or more characters after the string. The main target of this review is to check for all the existing research works to predict crime. In addition, we want to help the research community by identifying the different datasets used to apply the algorithms. Irrelevant studies are removed by applying multiple filters to our search queries. We also selected 30 papers to be part of the main text based on relevance and novelty, and 20 more papers are added in the appendix section \ref{tab a1}. In this survey, we have used a combination of an automated and manual search shown in Figure \ref{fig1}. In the initial stages, we focused on using the automatic digital search. In the final step, we manually read the entire paper to select a set of documents in machine learning and deep learning areas. 

Firstly, we have identified the key terms to create the queries. We then used those keywords to construct the various research database-related queries based on respective syntax. Below are the queries used to explore IEEE, Science Direct, and ACM databases.

IEEE  query:

((``Document Title'': ``crime*") AND (``Document Title": ``predic*" OR ``Document Title": ``detec*" OR ``Document Title": ``recogni*" OR ``Document Title": ``machine learning" OR ``Document Title":``deep learning"  OR ``Document Title":``clustering" OR ``Document Title":``natural language processing")) 

Science Direct Query:

(``crime") AND ( ``prediction" OR ``detection" OR ``recognition" OR ``machine learning" OR ``deep learning" OR ``clustering" OR ``natural language processing")

ACM Query:

``query": \{ Fulltext:((``crime*") AND ( ``predic*" 
OR ``detec*" OR ``recogni*" ``machine learning" OR ``deep 
learning" OR ``clustering" OR ``natural language processing")) \}

\subsection{Data Collection}\label{subsec2}

We focused on looking into individual research libraries rather than searching in google scholar. Because google scholar will have data from all these databases, there could be duplicates. Below are the database library homepage links where the research works were extracted using the keywords mentioned in search queries. Initially, we searched using all the metadata attributes available on each database. Next, we applied the filters only on the full-text papers. As we noticed that the number of documents is still high, we applied the filter on the index terms used in the article as the results will be more relevant. We have more than 450 papers from all the databases at this stage. Finally, in the last step, we applied the filter on the document title, where the total number of papers was 157.

a) Science-Direct Elsevier (https://www.sciencedirect. 
com) 

b) ACM (DL) Digital Library (https://dl.acm.org)

c) IEEE Xplore Access Digital Library (https://ieeexplore. 
ieee.org) 

\begin{figure}[h]%
\centering
\includegraphics[width=1\textwidth,height=1.25in]{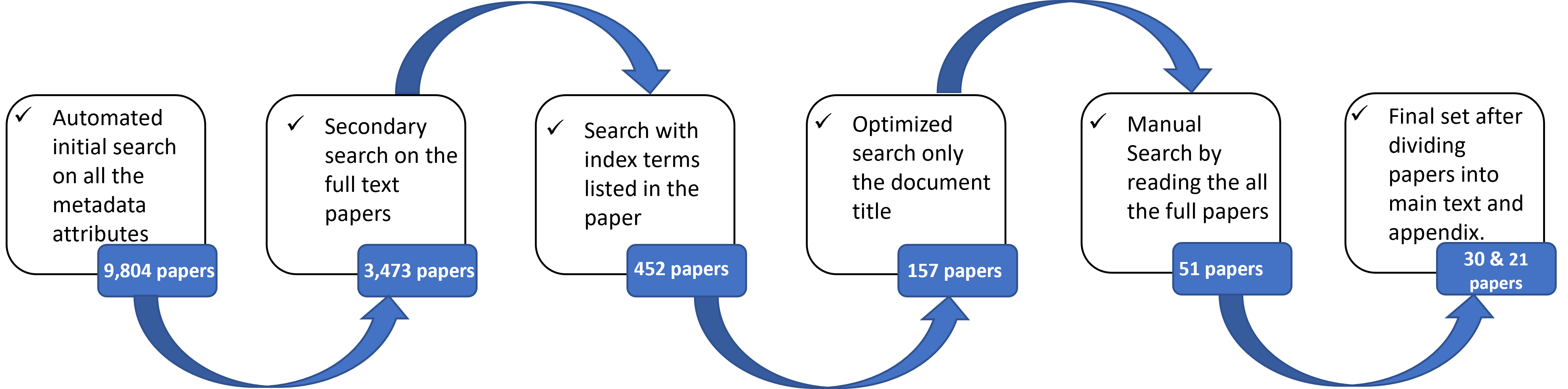}
\caption{Research paper selection methodology}\label{fig2}
\end{figure}

After applying the automated filters as shown in Figure \ref{fig2}, all the authors manually divided the work and read the papers to select the final set. For the selection, we mainly looked into the essential elements like the focus or objective of the article, datasets that authors have used, algorithms applied, and the accuracy rates. The focus of this survey is not only to help the community know the various algorithms applied but also to let them know about the datasets they can use to apply the novel algorithms and get the results for their research. 

\section{Literature Analysis}

\subsection{Pre-assessment Literature Analysis}\label{subsec2}

An analysis of collected literature data from the distinct research databases is essential \cite{p2} to receive information regarding the growth of an adopted research domain, scope across the research community, and popularity among the existing researchers. Thus, we have performed a detailed analysis of the collected literature data. We did pre-analysis and post-analysis, wherein pre-analysis comprises the exploration of initially collected literature ( i.e., the research papers that were collected immediately after performing our search query), and post-analysis comprises the investigation of those study’s data that were finally selected after applying selection criteria. Although we have exhumed the two renowned research databases, IEEE and Science Direct, lately, it has been observed that Science direct discontinued the search result extraction. Hence following analytical charts are based on IEEE databases findings.  
\begin{figure}[h]%
\centering
\includegraphics[width=0.9\textwidth,height=1.9in]{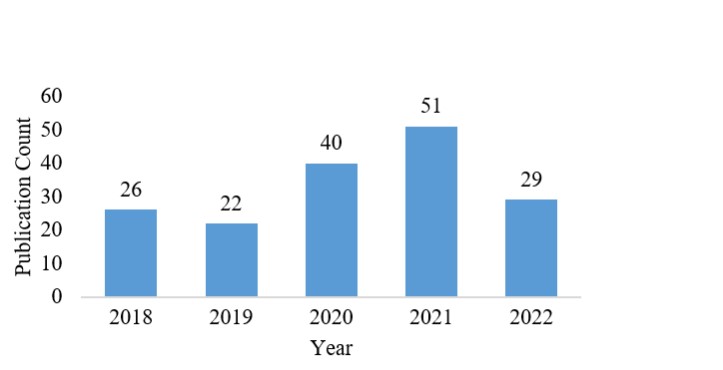}
\caption{Research publication trends from 2018-2022}\label{fig3}
\end{figure}
Figure \ref{fig3} shows the research publication trends in neighborhood crime from 2018 to 2022 (i.e., the last five years). It showed upward trends from 2018 to 2021 and downward trends during 2022. The years 2020 and 2021 are the apexes of COVID-19, which could be why existing researchers have utilized that time to explore more neighborhood crime research. 

\begin{figure}[h]%
\centering
\includegraphics[width=.9\textwidth,height=1.7in]{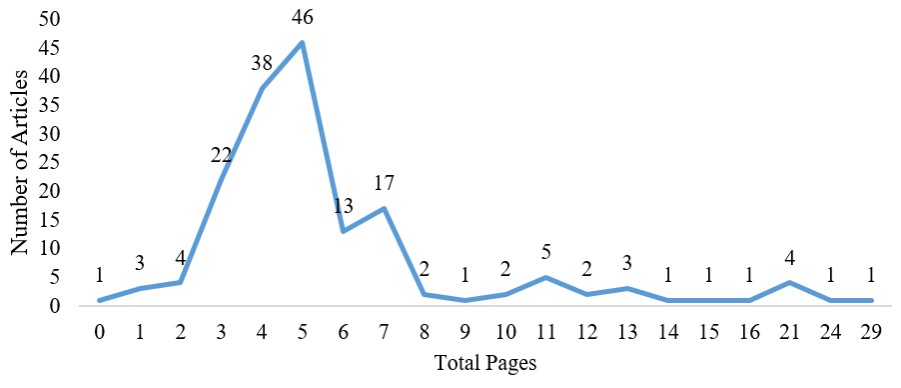}
\caption{Distribution of article's page counts}\label{fig4}
\end{figure}

Figure \ref{fig4} indicates the distribution of article page counts for the neighborhood crime research. The majority of articles (i.e., 46) comprise five pages. This may be because many researchers have published their work in conferences and symposiums rather than journals. Figure \ref{fig4} depicts that very few articles comprise more than ten pages. Since conferences, seminars, workshops, and symposiums allowed the presentation of abstracts and short papers, this Figure \ref{fig4} also depicts the 1,2,3 page-long articles.  

\begin{figure}[h]%
\centering
\includegraphics[width=0.90\textwidth,height=2.25in]{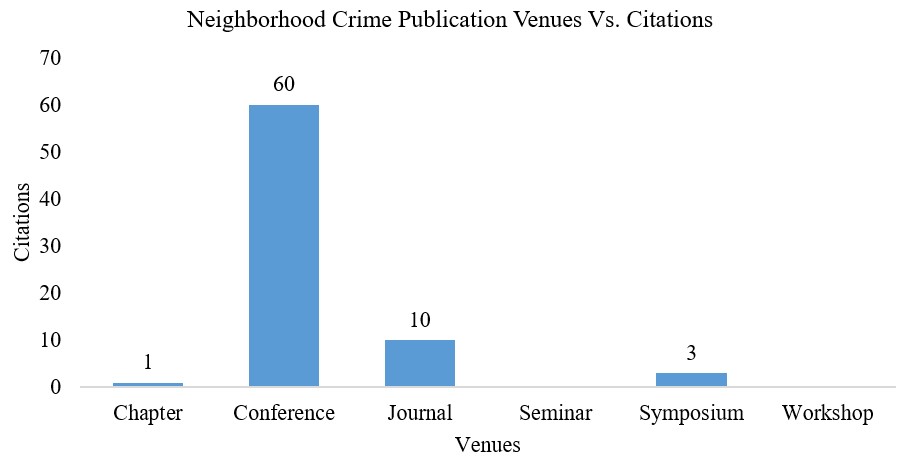}
\caption{Distribution of article's citations}\label{fig5}
\end{figure}

To assess the popularity of the neighborhood crime research articles, we have also performed the citation analysis shown in Figure \ref{fig5}. Wherein the conference papers have gained more citations (60) than the other journals (10 citations), chapter (1 citation), seminar (0 citations), symposium (3 citations), and workshop (o citation) articles in the neighborhood crime area, this may be because this area is less popular among the researchers. 
\begin{figure}[h]%
\centering
\includegraphics[width=1\textwidth,height=3in]{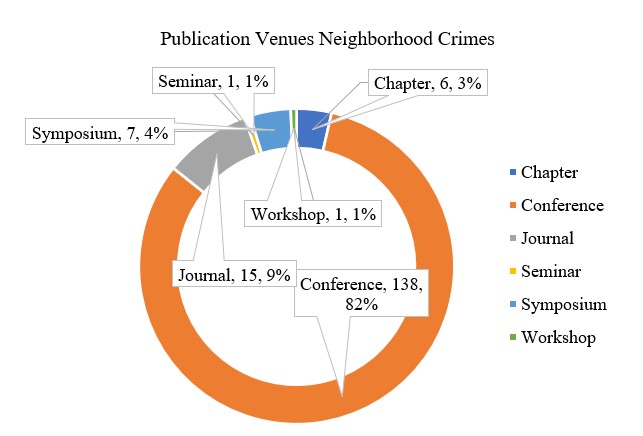}
\caption{Distribution of research articles at various venues}\label{fig6}
\end{figure}
Figure \ref{fig6}, shows the distribution of neighborhood crime-related articles published in various venues. Wherein this figure depicts the venue name (e.g., conference, journal, etcetera), number of papers (i..e, in numeric value-1,7..), and percentage of published articles (i.e., 1\%, 82\%,..). Here, the majority of papers have been published in the conferences than the other venues like journals, workshops, symposiums, and seminars. In the neighborhood crime area, 82\% of the articles were published in conferences, followed by 9\% in the journal, 4\% in the symposium, 3\% in book chapters,1\% in seminars, and 1\% in workshops.

\subsection{Post-assessment Literature Analysis}\label{subsec3}
To enhance the understanding of the neighborhood crime domain in combination with our above-cited selection criteria, we have also performed the post-assessment literature analysis because this is the crux of our literature survey to fulfill our identified objectives. We have created a word cloud for Neighborhood crime-related papers to fathom further the selected papers' underlying key concepts or themes. A word cloud often called a tag cloud, is a graphic depiction of the terms that appear the most frequently in a given text. Each word's magnitude in the word cloud reflects how frequently it appears in the text. Word clouds are frequently employed in literature reviews to swiftly pinpoint the key themes or topics within a sizable body of material. Additionally, they can be used to compare various texts and find trends and patterns in the data ~\cite{p1}
\begin{figure}[h]%
\centering
\includegraphics[width=0.85\textwidth,height=2.25in]{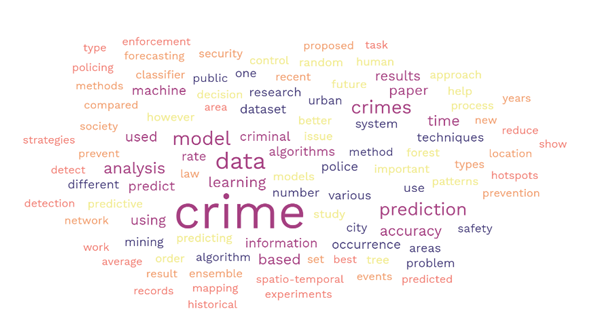}
\caption{Word cloud on selected articles}\label{fig7}
\end{figure}

\begin{figure}[h]%
\centering
\includegraphics[width=1\textwidth,height=3in]{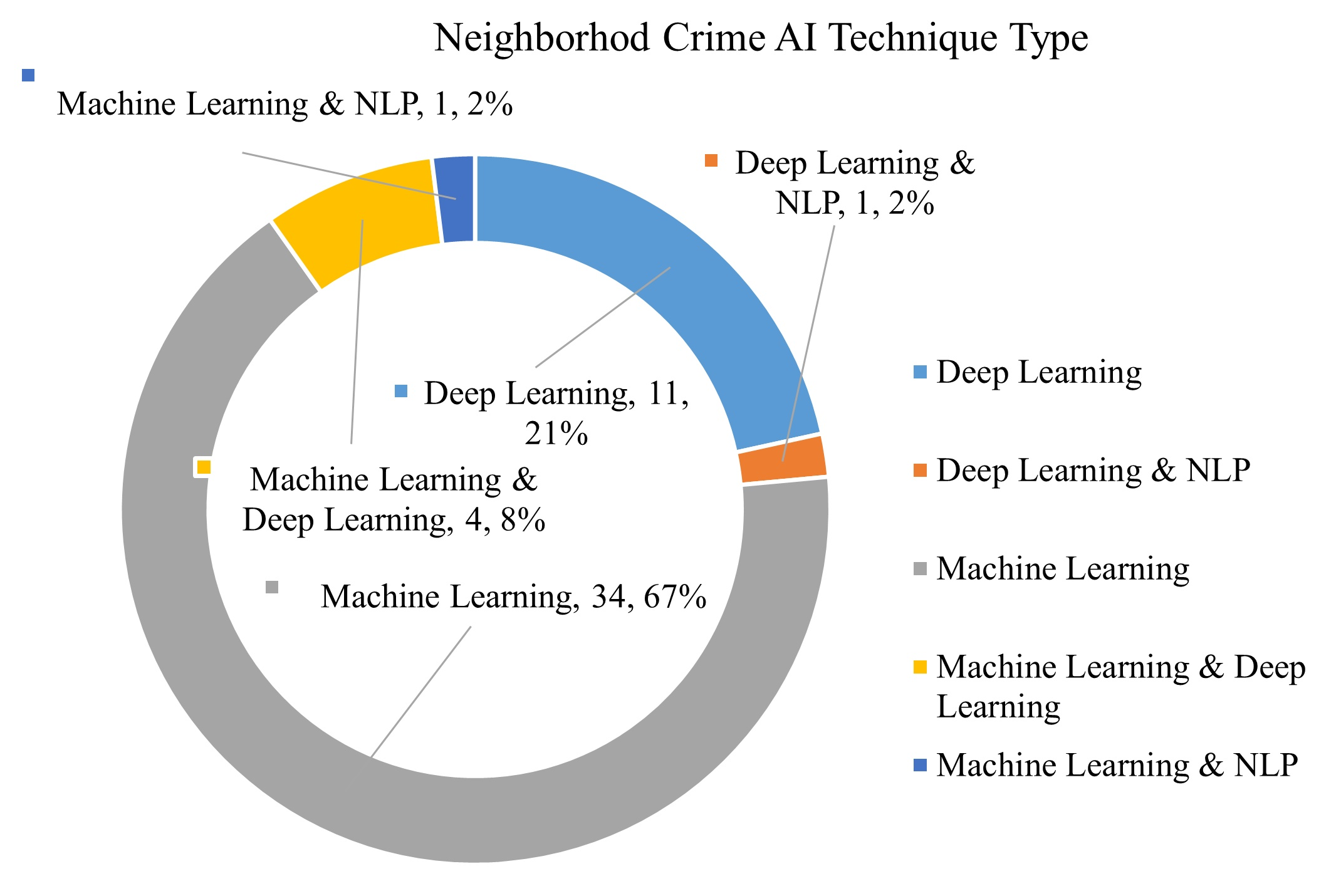}
\caption{Distribution of neighborhood crime-related selected article’s technique types}\label{fig8}
\end{figure}
Figure \ref{fig7} shows the word cloud for the neighborhood crime-related selected articles wherein many crime-related key terms have emerged as trends in the existing studies. “crime”, “criminal”, “policing”, “enforcement”, and “security” are the words that indicate the emphasis of researchers on the sub-areas of crime detection. Moreover, “prediction”, "algorithms", and “techniques” are the words that indicate the aim of the existing studies. This word cloud is aligned with our selection criteria and objectives that further validate our final set of selected articles for this literature review study.

Figure \ref{fig8} shows the Distribution of neighborhood crime-related selected article’s technique types among Science Direct and IEEE databases, respectively.  As shown in the Figure \ref{fig8}, among the neighborhood crime-related articles, machine learning (ML), the combination of machine learning and deep learning (DL), and the combination of machine learning and natural language processing (NLP), the combination of DL and NLP, and DL are the majorly used AI technique type. In the neighborhood crime domain – 67\% of articles have used ML, 21\% have used DL, 8\% have used the combination of ML and DL, 2\% have used the combination of DL and NLP, and 2\% have used the combination of ML and NLP. It has been observed that machine learning techniques are popular in the neighborhood crime area.

\begin{figure}[h]%
\centering
\includegraphics[width=1.1\textwidth,height=2.50in]{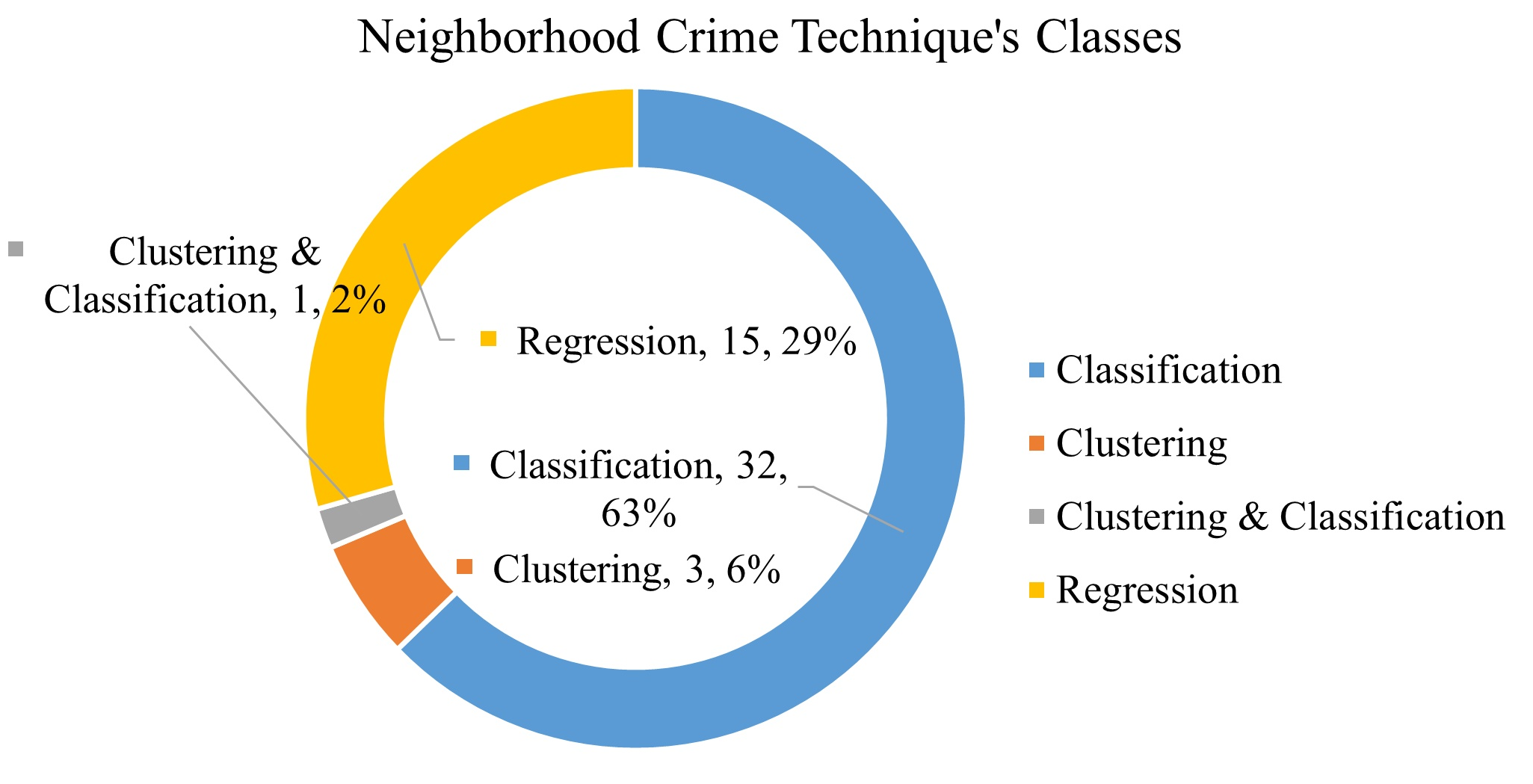}
\caption{distribution of neighborhood crime-related selected article’s technique’s classes}\label{fig9}
\end{figure}

Figure \ref{fig9} shows the distribution of neighborhood crime-related selected article’s technique classes among Science Direct and IEEE databases. As shown in Figure \ref{Fig. 9}, among the neighborhood crime-related articles, the classification task is the prime focus utilizing various ML and DL-related techniques. In the neighborhood crime domain – 63\% of articles have focused on the classification task, 29\% have focused on the regression tasks, 6\% have focused on the clustering task, and 2\% of the studies have utilized the combination of classification and clustering. It has been observed that the Classification task is the prime focus of the studies on neighborhood crime.  

\begin{figure}[h]%
\centering
\includegraphics[width=1\textwidth,height=2.50in]{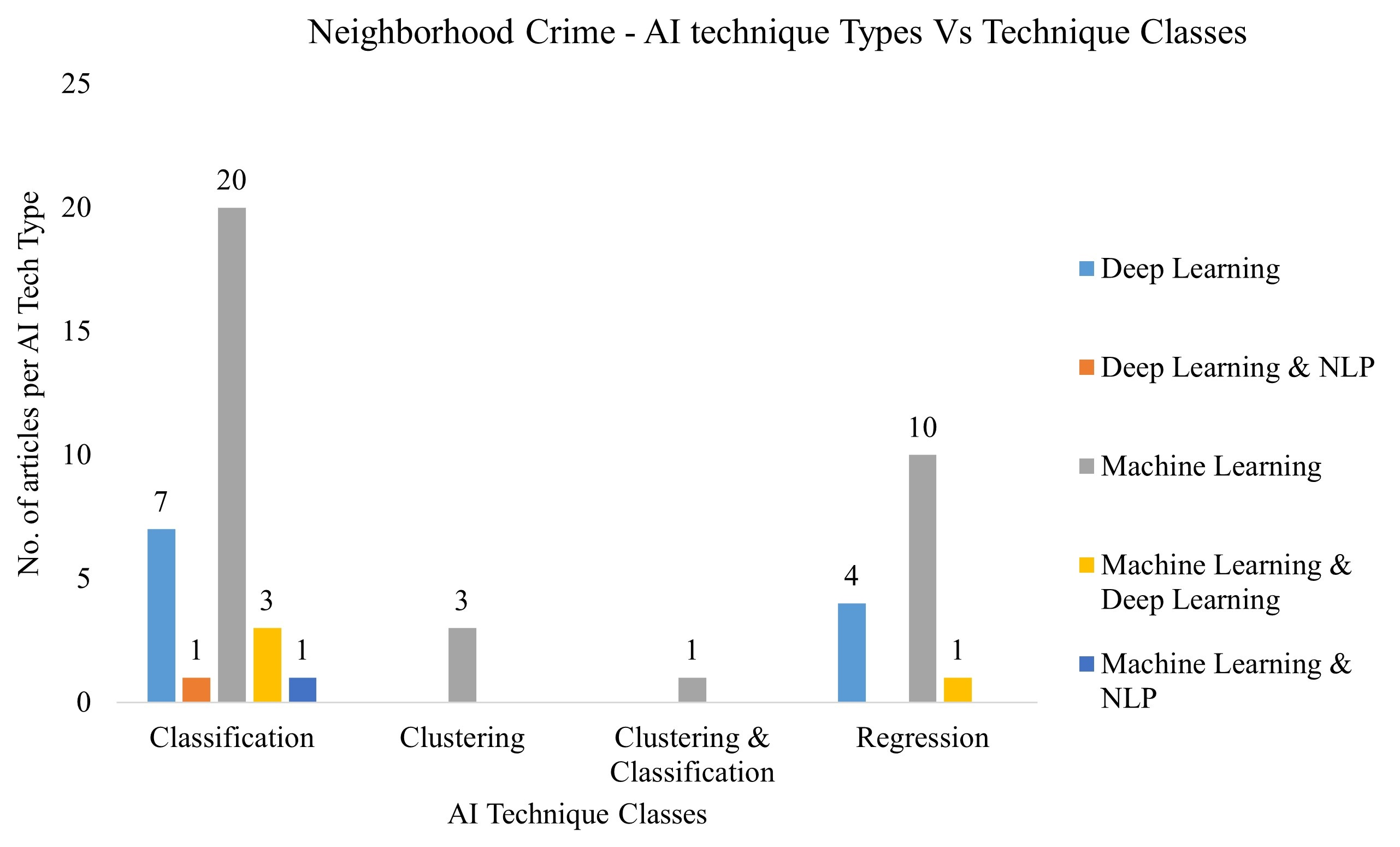}
\caption{Distribution of neighborhood Crime related selected article’s Technique classes Versus technique Type}\label{fig10}
\end{figure}
Figure \ref{fig10} shows the Distribution of neighborhood Crime related selected article’s Technique classes Versus technique Type among Science Direct and IEEE databases. Figure \ref{fig10} answers the question- what AI Technique is used for which technique classes (classification, clustering, and regression)?  Herein all five techniques (ML, DL, DL+NLP, ML+NLP, and ML+DL) have been used for the classification task whereas, for clustering and the combination of clustering and classification, the ML is solely used. In neighborhood crime articles NLP is also used for classification and regression tasks in addition to ML and DL. For the classification and regression tasks ML and DL both have been used.

\section{Crime Prediction Process \& Datasets} \label{sec3}

\begin{figure}[H]%
\centering
\includegraphics[width=1\textwidth,height=1.7in]{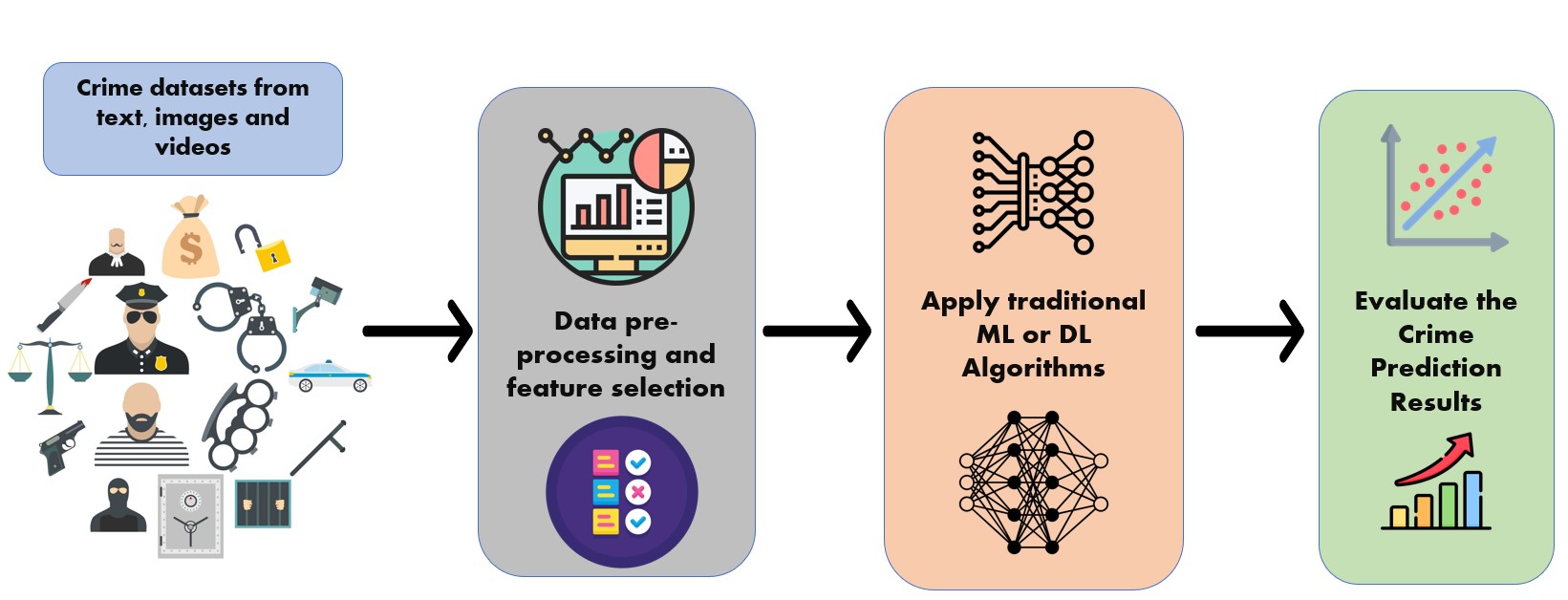}
\caption{Architecture flow of crime prediction}\label{fig11}
\end{figure}

Crime prediction using machine and deep learning involves several major steps as shown in Figure \ref{fig11}. The first step is data collection, which involves gathering relevant data such as crime statistics, demographics, and weather patterns. The next step is data preprocessing, which includes cleaning and transforming the data into a usable format. After data preprocessing, the data is split into training and testing sets for model development and evaluation. The next step is feature engineering, which involves selecting relevant features from the data that can be used to train the model. Once the features are selected, various machine and deep learning algorithms can be applied to the data for training and prediction purposes. Finally, the trained models are evaluated using various performance metrics to assess their accuracy and effectiveness in predicting crime. The results can be used to support decision-making in law enforcement and crime prevention efforts.

\begin{longtable}{L{0.6\textwidth}L{0.15\textwidth}
   L{0.2\textwidth}}
    \caption{Crime detection papers with data science algorithms}\label{tab1}%
        \\ \hline
        Dataset Links & Type of Data & Country - Locality \\ \hline
        https://data.gov.in/resource/incidence-crime-committed-against-women-india-during-2001-2012 \cite{tamilarasi2020diagnosis} & Crime & India \\ \hline
        https://indore.mppolice.gov.in/ \cite{kumar2020crime} & Crime & India - Madhya Pradesh \\ \hline
        https://ncrb.gov.in/en/node/3721 \cite{agarwal2018crime}\cite{kshatri2021empirical}\cite{kumar2020crime}\cite{bandekar2020design}\cite{sangani2019crime} & Crime & India \\ \hline
        https://data.gov.in/catalogs \cite{agarwal2018crime}\cite{sivanagaleela2019crime} & Crime & India \\ \hline https://www.kaggle.com/datasets
        /murderaccountability/homicide-reports \cite{shermila2018crime}& Crime & Unites States \\ \hline
        https://geodash.vpd.ca/opendata/ \cite{kim2018crime} & Crime & Canada - Vancouver \\ \hline
        https://plenar.io/explore/ \cite{catlett2018data}\cite{catlett2019spatio} & Crime - Spatio Temporal & Unites States - Chicago \\ \hline
        https://data.cityofchicago.org/ \cite{yi2018integrated}\cite{dash2018spatio} \cite{han2020risk}\cite{li2022spatial}\cite{tasnim2022novel}\cite{zhou2022unsupervised}\cite{butt2021spatio} & Crime  & Unites States - Chicago \\ \hline
        https://datasf.org/opendata/\cite{yao2020prediction} \cite{tasnim2022novel} & Crime & United States - San Francisco \\ \hline
        https://data.london.gov.uk/dataset/ \cite{sathiyanarayanan2019visual}& Crime & United Kingdom - London \\ \hline
        https://br-city.survey.okfn.org/place/rn.html \cite{araujo2018towards}& Crime & Brazil - Natal \\ \hline
        https://opendata.cityofnewyork.us/data/ \cite{butt2021spatio}\cite{almuhanna2021prediction}\cite{li2022spatial}\cite{zhou2022unsupervised}\cite{baqir2020evaluating}\cite{catlett2019spatio}\cite{elluri2019developing}& Crime  & United States - Newyork \\ \hline
        www.twitter.com \cite{algefes2022text}\cite{sandagiri2020detecting}\cite{permana2021crime} & Social Media & Global - Twitter \\ \hline
        https://www.kaggle.com/competitions
        /sf-crime/data \cite{sangani2019crime}& Crime & United States - San Francisco \\ \hline
        https://data.lacity.org/ \cite{zhou2021mixed}\cite{zhou2022unsupervised}& Crime & United States - Los Angeles \\ \hline
        https://dasci.es/transferencia/open-data/24705/ \cite{shenoy2022intelligent}& Crime- Weapons & Global \\ \hline
        https://www.gdeltproject.org \cite{aldossari2022data}& Crime & Global \\ \hline
        www.foursquare.com \cite{zhou2022unsupervised}& Geospatial & Unites States \\ \hline
        https://www.nyc.gov/taxi \cite{zhou2022unsupervised}& Taxi & United States - Newyork \\ \hline
        https://www.wunderground.com/weather
        /api \cite{zhou2022unsupervised}& Weather & Global \\ \hline      https://www.icpsr.umich.edu/web
        /NACJD/studies/3355/datadocumentation \cite{ma2022eadtc}& Crime & United States \\ \hline
        https://roc-ng.github.io/XD-Violence/  \cite{boukabous2022multimodal}& Crime - Video & Global \\ \hline
        www.facebook.com \cite{permana2021crime}& Social Media & Global - Facebook \\ \hline
        https://datasus.saude.gov.br/ \cite{alves2018crime}& Health & Brazil \\ \hline
        https://www.phillypolice.com/crime-maps-stats/ \cite{he2021prediction}& Crime - Spatio Temporal & Unites States - Philadelphia \\ \hline
        https://mapstyle.withgoogle.com/ \cite{he2021prediction} & Geospatial & Global \\ \hline
        https://www.mapbox.com/ \cite{he2021prediction} & Geospatial & Global \\ \hline
        https://data.police.uk/data/ \cite{toppireddy2018crime} & Crime & United Kingdom \\ \hline
        https://polisen.se/en/services-and-permits/police-record-extracts/ \cite{wolf2018prediction}& Crime & Sweden \\ \hline        https://www.openu.ac.il/home/hassner
        /data/violentflows/ \cite{sahay2022real} & Crime - Video & Global \\ \hline

\end{longtable} 

As shown in Table \ref{tab1}, there have been many datasets used in crime detection and prediction research articles. One example is the Chicago Crime Dataset, which contains data on crimes reported in the Chicago area. This dataset has been used to create models that predict the likelihood of specific types of crimes occurring in different areas of the city. Another dataset used in crime prediction research is the London Crime Dataset, which contains data on crimes reported in London city. This dataset has been used to create models that predict the likelihood of crimes occurring in specific areas and their relationship to the socio-economic factors of people based on their geo-locations in the area. 

Other datasets commonly used in crime detection and prediction research include the Los Angeles Crime Dataset, the New York City (NYC) Crime Dataset, and the Philadelphia Crime Dataset. These datasets contain information on crimes reported in their respective cities and have been used to create models that predict the likelihood of specific types of crimes occurring in different areas. In addition to these, there are also global datasets that focus on CCTV video footage, types of aggression, and weapons for real-time crime predictions.  

Overall, these datasets provide valuable information for researchers to build crime prediction models that could help law enforcement agencies prevent and respond to criminal activities more effectively. The location and access to datasets used by research articles surveyed in this paper are listed in Table \ref{tab1}.

\section{Crime Prediction using Machine Learning techniques}\label{subsec4}
Traditional machine learning models have proven to be effective for crime prediction. Various types of models such as decision trees, support vector machines, logistic regression, and random forests have been utilized to analyze crime data and identify patterns that can be used to predict criminal activity. Unlike deep learning, which relies on large amounts of data and complex neural networks, traditional machine learning models require fewer data points and are easier to interpret. For example, a logistic regression model can be used to predict the likelihood of a certain type of crime occurring based on factors such as time of day, location, and demographics of the area. A decision tree model can be used to identify the most important factors that contribute to the occurrence of a particular crime. Random Forest (RF) models can be used to analyze a wide range of features and make predictions about crime patterns. In addition to these techniques, traditional machine learning models can also be used for anomaly detection and outlier analysis in crime data. By identifying unusual patterns or outliers in the data, law enforcement agencies can detect potential criminal activity and take action to prevent it. In the below sections 5.0.1 and 5.0.2, we discuss the latest research on using machine learning model-based regression and classification for crime prediction.

\subsubsection{Machine Learning based Regression Methods for Crime Prediction}\label{subsec4}

Several crime detection scenarios are predicted using regression techniques as shown in \ref{tab2}. Researchers mainly focused on prevalent crimes like motorcycle robbery, losing property, and crimes in urban areas. Numerous factors may drive the boom in motorcycle robberies. For example, population growth and density, commuting conduct, bike usage, etc. These situations are problems for the police to govern and screen regularly since it requires forecasting and probabilities of robbery in a precise term. A novel method is proposed in the research study \cite{utomo2018prediction}; the authors created an application to predict motorcycle robbery with a technique to consider outside consequences using ARIMAX – TFM with a single input. The accuracy of ARIMAX is measured using Mean Absolute Percentage Error (MAPE) and Root Mean Squared Error (RMSE), and the scores are 32.30 and 6.68.

Rapid urbanization is a compelling challenge connected to city management and services. Cites with higher crime rates are difficult to manage public safety. To reduce crimes, new technologies are relieving police departments to access vast amounts of crime data to identify underlying trends and patterns. These technologies have doubtlessly grown the efficient deployment of police assets within a given region and ultimately guide greater powerful crime prevention. Researchers have worked on predictive models to use these datasets and predict crimes. Study \cite{catlett2019spatio},  provides a technique primarily based on spatial analysis and auto-regressive fashions to automatically locate excessive-hazard crime areas in city areas and reliably forecast crime tendencies in each area. Experiments are performed on real-world datasets gathered in New York City and Chicago. 

Another study \cite{ingilevich2018crime}, compared multiple techniques to predict the crimes in various areas of a metropolis. This research explored three predictive models: linear regression, logistic regression, and gradient boosting. The authors utilized feature selection techniques to select essential predictors. By applying feature selection methods, there is an improvement in accuracy scored, and it helped to avoid model overfitting. After comparing the results of all four models, the authors found that the gradient boosting technique outperformed, proven to be the best method to predict the crime rate in the urban area. 

In another study \cite{da2020prediction}, authors have looked at the crimes in Brazil, which have increased rapidly in recent times. Numerous predictive solutions use intelligent systems to identify when will a criminal offense will arise, which lets police send to those areas that are in danger. As part of their research, the authors looked into four machine-learning approaches for identifying where a criminal offense will arise in Fortaleza, Brazil. Their results indicate that easy algorithms are efficient in predicting crime. Also, they have seen that the Decision Tree and Bagging Regressor strategies obtained quality prediction outcomes. 

As mentioned above, numerous linear models are there to predict crime through the correlations between urban metrics and crime. However, due to multicollinearity and nonGaussian distributions in urban attributes, we usually tend to make controversial conclusions on these attributes to predict crime. Ensemble-based machine learning algorithms can deal with such problems adequately. In the research work \cite{alves2018crime}, authors applied random forest regressor to predict the crime and quantify the impact of urban attributes on homicides. Their approach has 97\% accuracy in crime prediction, and the significance of city indicators is clustered and ranked equally. Their research identifies the rank of urban indicators based on their significance in predicting crime. As per their results, unemployment and illiteracy are the essential variables for depicting homicides in Brazilian towns. 

\begin{longtable}{L{0.4\textwidth}L{0.15\textwidth}
   L{0.15\textwidth}L{0.175\textwidth}}
    \caption{Crime detection using machine learning regression techniques}\label{tab2}%
    \\ \hline

        Methodology  & ML  Algorithms  & Dataset  & Performance  \\ \hline
        ARIMAX method is used to predict the time series data from theft cases that affected the number of motorcycles. \cite{utomo2018prediction}  & ARIMAX algorithm  & City of Yogyakarta  &  RMSE - 6.68  \\ \hline
        Crime data is used to detect and map crime-dense region (CDR) in NYC and Chicago, then ARIMA is applied on these crime hotspots. \cite{catlett2019spatio}  & ARIMA, Random Forest (RF), RepTree and ZeroR  & Chicago Crimes - 2001 to present,  New York City Crimes:, 2006 to 2016  & 2016 RMSE for Area - 143.73, CDR 1 - 57.8, CDR2 - 29.85 \& CDR 3 - 16.19   \\ \hline
        Clustering is used to understand distribution across city and then cluster are used to predict the number of crimes per location. \cite{ingilevich2018crime}  & Linear regression (LR), LoR and Gradient boosting.  & Saint-Petersburg Russia Crime (2014 - 2017)    & R-Square - 0.9  \\ \hline
  
        Four machine learning methods are applied to predict the location of where a crime, more specifically, where a crime of robbery will occur. \cite{da2020prediction} & RF, Bagging, DT, Extra tree & City of Fortaleza data & RMSE - 0.00231 \\ \hline
        Random forest regressor is used to predict crime and quantify the influence of urban. Results showed that unemployment and literacy are the two major indicators of crimes. \cite{alves2018crime} & Random Forest Regressor & Department of Informatics of the Brazilian Public Health System — DATASUS.  & Upto 97\% Accuracy, Adjusted R square 80\% on average. \\ \hline

\end{longtable}

\subsubsection{Machine Learning based Classification Methods for Crime Prediction}\label{subsec4}

Traditional regression techniques can successfully check the variables' significance but, they must be more reliable for crime prediction. In many research works \cite{sathiyanarayanan2019visual}\cite{araujo2018towards}\cite{almuhanna2021prediction} mentioned in section \ref{subsec4}, authors have proven that machine learning models effectively predict crimes. Still, they could be more efficient in identifying which variables are significant in predicting crimes. We further examined the classification techniques to predict different criminal incidents like analyzing the criminal reports as shown in Table \ref{tab3}. Studying those reviews for crime prediction enables regulatory authorities to deal with crime prevention strategies. However, collecting these reviews personally and determining their crime types is challenging. In one study \cite{das2021incremental} authors have created a novel approach, an incremental classifier that learns the new data and dynamically predicts the results. In this research \cite{das2021incremental}, they have utilized the Bi-objective Particle Swarm Optimization technique to develop an efficient incremental classifier for dynamically classifying and predicting crime reports. Crime reports from various countries have been collected from online newspapers to measure the performance of their classifier. Also, they evaluated the results manually with unprejudiced police witness narrative crime reports. They tested their approach on four datasets to measure their model's statistical significance. 

 \begin{longtable}{L{0.4\textwidth}L{0.15\textwidth}
   L{0.15\textwidth}L{0.175\textwidth}}
    \caption{Crime detection using machine learning classification techniques}\label{tab3}%
        \\ \hline
        Methodology  & ML  Algorithms  & Dataset  & Performance  \\  \hline
        OVR-XGBoost: One Vs Rest OVO-XGBoost: One Vs One. The primary difference is based on how the dataset is organized for classification. \cite{yan2022research}  & XGBoost based algorithms  & H City China: (2019)  & Aggregate Accuracy - 85\%  \\ \hline
        Applying particle swarm optimization-based classifier and a rules engine to classify crime reports. \cite{das2021incremental}  & Multiple classification algorithms  & USA, India \& UAE crime articles (2007 - 2017)  & Aggregate Accuracy - 79\%  \\ \hline
        17 spatiotemporal variables were used to train and test the XGBoost algorithm, and then SHAP is used to explain the model predictions. \cite{zhang2022interpretable}  & XGboost model  & ZG City, in Southeast China. (2017 to 2020 ).   & Accuracy - 89\% \& AUC - 0.586  \\ \hline
        Random under and over sampling techniques are used to evaluate multiple algorithms. \cite{trinhammer2022predicting}  & Logistic Regression (LoR), RF, XGBoost and LightGBM  & Danish national psychiatric patient register, (474 crime).   & F1 Score - 76\%  \\ \hline
        Graph theory based. Decision trees and heuristics are used for feature selection and an ensemble classifier is applied. \cite{das2022graph}   & Multiple classification algorithms  & USA, UAE \& India Crime Articles (2008 - 2016)  & Aggregate F1 Score - 88\% \\ \hline 
        Identify if the crime is spread uniformly according to population density or whether specific socio-economic attributes account for increased or decreased crime. \cite{sathiyanarayanan2019visual} & Visual Analysis & London crime dataset & Aggregate Accuracy - 97\% \\ \hline
        Follows an algorithm-as-a-service architecture (AaaS), providing insights into existing public safety systems and platforms. \cite{araujo2018towards} & K Nearest Neighbor (KNN) & Crime endogenous data sources & Aggregate Accuracy - 90\% \\ \hline
        Prediction of crime in neighborhoods of New York city using spatial sata analysis. \cite{almuhanna2021prediction} & XGBOOST, RF and SVM & New York crime data & Aggregate Accuracy - 52\% \\ \hline
        EADT approach is used for Interpretable and Accurate Crime Prediction. \cite{ma2022eadtc} & Decision Tree (DT) & Pennsylvania state prisons & Aggregate Accuracy - 77.6\% \\ \hline
        Traditional data science steps are used to cluster crimes and then predict using classification modeling methods. \cite{bandekar2020design} & K-Means & National Crime Records Bureau (India) & Aggregate Accuracy - 78\% \\\hline

\end{longtable}

Another research \cite{yan2022research}, focuses on predicting crime using the XGBoost algorithm.  Based on the records of theft instances in H city, they developed an optimized decomposition and fusion method based on XGBoost and applied multi-class classification models like OVR-XGBoost and OVO-XGBoost. As the theft datasets have different classes, they have utilized the SMOTENN algorithm to process and make data the dataset balanced. Their results show that OVR-XGBoost and OVO-XGBoost models' prediction accuracy is better than the baseline XGBoost models. In the study \cite{zhang2022interpretable}, the authors have selected 17 variables for crime prediction, and the XGBoost algorithm is adopted to train the prediction model. A post hoc interpretable approach, Shapley additive explanation (SHAP), is used to parent the contribution of person variables. SHAP, a post hoc interpretable method, is used to determine the significance of individual variables. Among all 17 variables used in this research, the percentage of the non-neighborhood population and the populace aged 25–44 contribute greater than different variables in predicting crime. The higher the ambient population of elderly 25–44 in the vicinity, the more public crimes. The authors have also validated the SHAP values to demonstrate each variable's contribution to the crime prediction across the experimental findings. These outcomes of the neighborhood techniques can assist the police in identifying the most important factors.At the same time, the global model identifies the essential features of the entire region. 

 Another research \cite{trinhammer2022predicting} focuses on predicting crime during or after psychiatric care. As modern threat-evaluation equipment is time-consuming to administer and offers constrained accuracy, this research looked to expand a predictive model designed to discover psychiatric patients liable to commit the crime. The authors utilized the longitudinal nice of the affected Danish person registries, recognizing the 45.720 adult patients who had connected with the psychiatric system in 2014, of which 474 committed crimes leading to a forensic psychiatric treatment direction after discharge. Authors have used four gadget studying models (Random Forest, Logistic Regression, XGBoost, and LightGBM)  over various sociodemographic, judicial, and psychiatric variables. Their model identified 47\% of future forensic psychiatric patients, making correct predictions in 57\% of samples. This research demonstrates how a clinically useful preliminary risk assessment is achieved using machine learning classification techniques. Their research helps to flag possible forensic psychiatric patients while in contact with the general psychiatric system, which allows early intervention initiatives to be activated. 

Another research work \cite{das2022graph} presents a graph-based ensemble classification approach for predicting crime reports better than traditional classifiers. Crime reports are graphically modeled to locate the maximal independent subset of features, and then they use decision tree classifiers on this set.  Extensive experiments are performed to compare the overall performance of the proposed approach on numerous crime data sets. The developed ensemble classification model demonstrated better performance. Apart from predicting crime, researchers \cite{ma2022eadtc}\cite{boukabous2022multimodal} also focused on interpreting crime-related predictions. This will lead to a better understanding of what impacts crime detection.

\subsection{Crime Prediction using Deep Learning techniques}\label{subsec5}

Deep learning has become a popular method for crime prediction in recent years. The studies included in the reference research articles use a range of deep learning algorithms, such as convolution neural networks (CNN), deep neural networks, and sentiment analysis, to analyze various types of data, including text, images, audio, and social media. These algorithms are capable of detecting patterns and anomalies in the data that can indicate criminal activity. One of the key strengths of deep learning is its ability to handle large and complex datasets, making it well-suited to the task of crime prediction. For example, image analysis algorithms can detect threatening objects in crime scenes and predict the likelihood of a crime occurring. Text mining techniques can be used to analyze crime-related tweets and make predictions about crime patterns. In addition, deep learning algorithms can detect anomalies in crime data in smart cities, which could indicate the presence of criminal activity. Researchers used these techniques to tackle both regression and classification problems in crime prediction as detailed in the below sections.

\subsubsection{Deep Learning based Regression Methods for Crime Prediction }\label{subsec5}
Deep learning algorithms in regression analysis are used as a tool for crime prediction to identify the factors most strongly associated with crime and use these relationships to make predictions about future crime patterns. The research articles in this area highlight the strengths of regression in modeling the relationship between multiple variables, including crime data, weather data, demographic data, social media data, and location data. A common theme among these research articles shown in Table \ref{tab4} are the use of regression combined with deep learning techniques, such as convolution neural networks, recurrent neural networks, attention mechanisms, and sequential fusion models, to improve the accuracy of crime prediction.  

\begin{table}[!htbp]
    \centering  
    \setlength\tabcolsep{1pt}
    \caption{Crime detection using deep learning regression techniques}\label{tab4}%
    \small
    \begin{tabular}{ L{4cm}L{2.5cm}L{2cm}L{1.5cm}}
        \\ \hline
        Methodology  & DL Algorithm  & Dataset  & Performance  \\ \hline
        A Graph Convolution network in combination with ST-ResNet is used to perform spatiotemporal analysis and then LSTM is used to detect crimes in each community. GBDT is used to combine outputs of GCN and LSTM. \cite{han2020risk}  & Long Short Term Memory (LSTM) in combination with Spatio Temporal Graph Convolution Network (ST-GCN) and Gradient Boosting Tree  & Chicago Crime Data (2001 - 2020)  & R-Squared - 0.84  \\ \hline
        Developed Attention-LSTM to process categorical-temporal data and Stacked Bidirectional LSTM model to process spatial information. The two were fused using feature and decision-level fusion. \cite{tasnim2022novel}  & ATTN-LSTM, St-Bi-LSTM, Fusion Models  & San Francisco and Chicago Crime Data (2004 to 2017)  & Aggregated R-Squared $> 0.90$ \\ \hline
        LA Crime data is separated into block-wise information based on the hour of the day, area, and city. These blocks are used to train CNN model.\cite{zhou2021mixed}  & Mixed Spatio Temporal Neural Network based on CNN  & Los Angeles Crime Data  & RMSE - 0.22 \\ \hline
        A model framework based on three phases i.e., intercity similar-grid matching, auxiliary features construction, and crime risk prediction using a dense CNN-based unsupervised domain adaptation. \cite{zhou2022unsupervised} & CNN based on unsupervised domain adaptation model (UDAC)  & New York City, Chicago, and Los Angeles Crime, weather and taxi Data (2015)  & RMSE - 0.62  \\ \hline
        Weather, holiday, time slot ID, and Day of the week for spatial dependency and GRU-based features for Temporal dependencies to predict a number of incidents in different locations. \cite{liang2022towards}  & LoR, Support Vector Regressor, RF, GRU, Deep Crime and T-GCN  & Xiaogan, data (2016). New York City data (2014). Additionally POI, Urban anomaly and Weather data are used.  & Accuracy Xiaogan Dataset - 54\%, NYC Dataset - 61\% \\ \hline
    \end{tabular}
\end{table}

In research focusing on theft crime prediction \cite{han2020risk}, the authors use regression to model the relationship between theft crime data, demographic data, and weather data. The regression model adopts two deep learning models, a Long Short-Term Memory (LSTM) network and a Spatio-Temporal Graph Convolutional Network (ST-GCN), to predict the likelihood of theft crimes in urban communities. The regression model can incorporate external information, such as weather data, which can influence crime patterns. The LSTM and ST-GCN models capture the temporal and spatial dependencies in the data, respectively. In another article \cite{tasnim2022novel}, the authors use regression to model the relationship between crime data, weather data, and social media data. The regression model is part of a more comprehensive, multi-module approach that uses attention mechanisms and sequential fusion models to predict the likelihood of crimes. This framework consists of four sub-modules, where the initial two modules adopt St-BiLSTM and ATTN-LSTM to process temporal and spatial features. Finally, two fusion models are used to abstract the data and make crime predictions on Chicago and San Francisco crime datasets. 

In another research focused on using spatiotemporal data \cite{zhou2021mixed}, the authors use convolutional neural networks to develop a regression model on publicly available crime data in Los Angeles. The regression model is part of a more extensive, mixed spatiotemporal neural network that is designed to make real-time predictions about the likelihood of crimes. The authors claim that using regression in combination with the diverse spatiotemporal neural network results in improved accuracy and real-time performance. Another research \cite{zhou2022unsupervised} that applies crime risk prediction across different cities uses regression to model the relationship between crime data and demographic data from other cities. The regression model is part of an unsupervised domain adaptation technique designed to predict the likelihood of crimes in new cities. The authors claim that using regression in combination with the unsupervised domain adaptation technique results in improved accuracy in crime prediction. A recent research article \cite{liang2022towards} applied machine learning and deep learning methods to crime data from Xiaogan, a medium-sized city in China, to predict crime hourly. The models use weather, holiday, time slot ID, and Day of week information to extract spatial dependency (distance graph, poi similarity, and crime similarity). Temporal dependencies captured using GRU are used to predict the number of incidents in different locations.

These research articles highlight the versatility of regression as a tool that can be integrated with other techniques to enhance the performance of crime prediction models. Another commonality among the papers is the use of regression to model the relationship between crime data and other variables, such as weather and demographic data, to incorporate external information that may influence crime patterns. This allows for the creation of more comprehensive and accurate models of crime patterns. In summary, the five research papers demonstrate the strengths of regression as a tool for crime prediction, including its ability to model the relationship between multiple variables, its versatility in being integrated with other techniques, and its ability to incorporate external information that may influence crime patterns. 

\subsubsection{Deep Learning based Classification Methods for Crime Prediction}\label{subsec5}

Deep learning algorithms are trained on large amounts of data to classify instances into various categories. This makes them ideal for solving classification problems in crime detection. Deep learning models can accurately organize criminal activity and detect criminal intent by analyzing vast amounts of data, including images, audio, text, and social media data. For example, image-based data can provide detailed information about crime scenes, including the presence of weapons and other objects that may indicate criminal intent. Similarly, audio-based data can provide valuable insights into the tone and context of a conversation, helping to identify potential illegal activities. Another advantage of deep learning for classification problems in crime detection is the ability to identify hidden patterns in the data that traditional methods may miss. For example, deep neural networks can be trained to analyze crime-related tweets, uncovering patterns that indicate a potentially criminal act. The results of deep learning models in crime detection have been awe-inspiring. The papers reviewed under deep learning classification are listed in Table \ref{tab5}
 
\begin{table}[!htbp]
    \centering
    \setlength\tabcolsep{1pt}
    \caption{Crime detection using deep learning classification techniques}\label{tab5}%
    \begin{tabular}{ L{4cm}L{2.5cm}L{2cm}L{1.5cm}}
        \\ \hline
        Methodology  & DL Algorithm  & Dataset  & Performance  \\ \hline
        R-CNN algorithm is adopted detect multiple objects relevant to crime in real time videos and images. \cite{navalgund2018crime}  & VGGNET19 based on Fast RCNN \& RCNN  & Youtube and Google Videos \& Images  & Train Accuracy -  100\%  \\ \hline
        CNN is used to extract features from CCTV images to predict crime scene objects. \cite{nakib2018crime}  & CNN  & 2000 Images of weapons and blood scenes & Accuracy - 90.2\%   \\ \hline
        Classify object or person in video feed and track abnormal activity using Deep CNN (DCNN) and Recurrent Neural Network (RNN). \cite{chackravarthy2018intelligent} & DCNN and RNN  & 1 Million video streams [Karpathy]  & Mean Squared Error (MSE) - $1.38e-10$  \\ \hline
        Extract tweet data and process text to identify text keywords that are relevant to weapons used for crime or criminal activity. \cite{algefes2022text}  & CNN, KNN, NB, Support Vector Machine (SVM), DT and RF  & Saudi Arabia tweets (2017 - 2021).  & Accuracy - 79\% \\ \hline
        Live CCTV data is used to detect faces and weapons using CNN- GRU model. \cite{shenoy2022intelligent}  & RNN, Gradient Recurrent Unit (GRU) and LSTM.  & CAVIAR Dataset  & Accuracy - 95.97\%  \\ \hline
        Preprocess crime data to extract features and train both traditional and Deep Learning models to predict crime region and type. \cite{aldossari2022data}  & Traditional ML classifiers and Artificial Neural Network (ANN)  & Saudi Arabia (2018 - 2020)  & Accuracy for Crime Type - 99\% \& Crime Region - 81\%  \\ \hline
        Audio data is used to extract Mel-Frequency Cepstrum Coefficients from sound waves and a CNN-RNN classifier is deployed \& text data, features are extracted for BERT. Both models are combined by fusion model. \cite{boukabous2022multimodal}  & CNN \& BERT  & XD-Violence Audio \& Video Data  & Accuracy - 85.63\%  \\ \hline
        Twitter data is processed by keyword filtering and labeled. The features vectors are generated to be inputted into models for training. \cite{permana2021crime}  & SVM and ANN  & 100000 Tweets (January 2020)  & Accuracy - 90.3\%  \\ \hline
        Video data is used to extract spatio temporal features and gestures to train DRNN for classification (Hostility \& Violence). \cite{sahay2022real}  & Deep Reinforcement Neural Network (DRNN)  & Crowd Violence, UCSD and Violent Flow Data.  & Aggregate F1 score - 78\% \\ \hline
         Deep Learning and Machine learning models are applied to predict different types of crime and their relation to weather in Newyork city. \cite{elluri2019developing} & SVM, LR, DT, RF, CNN, RNN, LSTM and GRU & 2018 NYPD and Weather data & Accuracy - 99\% \\ \hline 
    \end{tabular}
\end{table}

In crime-related classification, two main types of deep learning algorithms are used: Convolutional Neural Networks (CNN) and Recurrent Neural Networks (RNN). CNN's are commonly used in image-based classification tasks, including crime scene prediction. In the research article focusing on crime scene data \cite{nakib2018crime}, CNNs are trained to detect threatening objects in crime scenes, such as weapons. This allows the model to comprehensively analyze the crime scene, including the presence of things that may indicate criminal intent. On the other hand, RNN's are commonly used to study temporal patterns in data. In a research article focusing on crime prediction based on behavioral tracking \cite{shenoy2022intelligent}, the authors use a combination of deep learning algorithms, including CNN and RNN, to analyze behavioral tracking data and motion analysis data. The study shows that this approach can effectively predict criminal activities, such as theft and robbery, by analyzing patterns in the behavior and movements of individuals in a given area. In the research articles \cite{algefes2022text} focusing on social media data, Artificial Neural Networks are trained on crime-related text data to predict the likelihood of a crime occurring. These models analyze the context and tone of the text data to classify patterns that may indicate criminal activity. 

In addition to the articles mentioned above, several other research studies \cite{navalgund2018crime}\cite{aldossari2022data}\cite{elluri2019developing} have also used deep learning techniques for crime prediction and classification. These studies demonstrate the versatility of deep learning algorithms in crime-related classification tasks, as they can be applied to a wide range of data types, including images, text, audio, and social media data. For example, research focusing on crime anomaly detection \cite{chackravarthy2018intelligent} uses deep learning algorithms, including Autoencoders and CNN, to analyze crime patterns in smart cities. The study shows that this approach can effectively detect unusual crime patterns, which may indicate the presence of criminal activity. In another study focusing on audio and text data \cite{boukabous2022multimodal}, the authors use a multimodal deep learning model based on CNN and BERT that considers both audio and text data to classify crime-related events. This model is advantageous when audio data, such as 911 calls, is available and can provide a complete picture of the crime event. 

These research articles including multiple other studies \cite{permana2021crime}\cite{sahay2022real} highlight that deep learning algorithms, including CNN and RNN, have successfully applied to various data types for crime prediction and classification. These studies demonstrate the versatility of deep learning algorithms in this field and provide valuable insights into the factors contributing to criminal activity. By leveraging the strengths of these models, law enforcement agencies can gain a more comprehensive understanding of criminal activity and take proactive measures to prevent crime from occurring. 

\section{Discussion and future work}\label{sec4}

The application of machine learning and deep learning algorithms to predict or detect crime has shown great promise in addressing this complex problem. By utilizing vast datasets and advanced algorithms, these technologies can potentially improve the accuracy and effectiveness of crime prediction models. However, despite the advances in this field, there are still significant gaps in our understanding of how these technologies can be effectively applied to the problem of crime prediction. In this section, we will discuss the potential benefits of machine learning and deep learning algorithms for crime prediction and the existing research gaps that need to be addressed in future research. 

One of the major advantages of machine learning and deep learning algorithms for crime prediction is the ability to analyze large datasets and identify patterns in criminal activity or behavior. The ability of these algorithms to process vast amounts of data, including social media and other online sources, can provide valuable insights into criminal activities that are yet to be committed. Furthermore, deep learning algorithms like CNN and RNNs have been used to analyze video footage from security cameras. This capability provides a more accurate and efficient means of detecting criminal activities. Another major benefit of machine learning and deep learning for crime prediction is the ability to develop real-time prediction models. These models can be used to analyze crime data in real-time and to predict future crime incidents. This supports law enforcement agencies to  act quickly if a criminal activity is being committed. Additionally, integrating decentralized machine learning algorithms with wearable technology, such as body cameras and smartwatches, provides new opportunities to collect and analyze data related to criminal activities. 

Even though machine learning and deep learning algorithms support effective crime prediction, there are still some significant challenges that need to be addressed. One of the major challenges in this area is the need for interpretable models that can provide clear explanations of their predictions. This is particularly important in the context of crime prediction, as incorrect predictions might lead to serious consequences for individuals and communities. Apart from the existing model-based explanation methods, it is also important to incorporate causal based explanations that focus on cause and effect relationship between crime patterns and relevant feature variables. Another challenge that needs to be addressed is the need for more accurate and reliable data. In order to effectively apply machine learning and deep learning for crime prediction, it is important to have access to high-quality and up-to-date crime data. As this study showed that many earlier researchers took advantage of data related to demographics, whether outside of crime-relevant datasets, there is a need to develop algorithms that can accurately handle data from multiple sources and integrate it into a single predictive model. 

Another significant area of focus should be to research more on the ethical implications of using machine learning and deep learning for crime prediction. As these technologies are used to predict individuals and communities, it is important to ensure that they do not perpetuate existing biases or lead to discrimination. Furthermore, there is a need for more research on the privacy implications of using these technologies for crime prediction, this included but not limited to the potential risks of data breaches and the misuse of personal information. Another significant gap in the existing research is the need for more research studies on the effectiveness of machine learning and deep learning for crime prediction in the real world. While these technologies have shown great promise in this area, there is a need for more rigorous evaluations of their accuracy and effectiveness in real-world scenarios. Additionally, there is a need for more research on the scalability of these technologies and the challenges associated with their implementation in large-scale systems. 

In conclusion, machine learning and deep learning methodologies have the potential to transform the field of crime prediction by providing more accurate and effective methods for predicting criminal activities. However, in order to fully realize the potential of these technologies, it is important to address the existing research gaps and challenges, including the need for interpretable models, accurate and reliable data, ethical considerations, and more rigorous evaluations of their accuracy and effectiveness. By addressing these gaps, we can advance our understanding of the role of machine learning and deep learning algorithms in crime prediction and contribute to the development of more effective and efficient policing strategies. 
\begin{table}[!htbp]
    \centering
    \setlength\tabcolsep{1pt}
    \caption{Future Research Directions}\label{tab06}%
    \begin{tabular}{ L{12cm}L{2.5cm}L{2cm}L{1.5cm}}
    \hline Concept Areas and Research Questions  \\ \hline
       
    \hline Feature-Oriented \\ \hline

1. How do visual features (e.g., images/videos) affect (i.e., positively or negatively) the crime identification?

2. How do audio features (e.g., voiceover, call recordings) affect (i.e., positively or negatively) the crime identification? 

3. How do Textual features (e.g., investigating notes, complaint statements, crime interviews) affect (i.e., positively or negatively) the crime identification?

4. What are the visual features to be considered to detect neighborhood crime?

5. What are the Audio features to be considered to detect neighborhood crime?

\\ \hline Technique-Oriented \\ \hline

1. Are there any reinforcement learning techniques available to detect neighborhood crime?

2. Are there any transfer learning techniques available to detect neighborhood crime? 

3. Are there any generative adversarial neural network techniques available to detect neighborhood crime?

\\ \hline 
    \end{tabular}
\end{table}

As a future research goal and agenda, we have illustrated a range of prospective research directions in the area of neighborhood crime based on the importance and also current lack of focus in the areas. From Table \ref{tab06}, future researchers may want to address concerns like - "Are there any reinforcement learning techniques available to detect the neighborhood crime?", and What are the visual features to be considered to detect the neighborhood crime? The relevant datasets are available for such identified future research questions; such data may be utilized to accomplish the goal of early recognition of neighborhood crime.  

The presented literature base and futuristic research goals offer a number of elements to direct future studies and thus theoretically support the effort to identify neighborhood crimes. Our systematic review offers a thorough grasp of the characteristics and methods used by earlier research to recognize and detect crimes. In addition, we have outlined 8 more research questions that fall into the categories of technique-oriented questions and feature-oriented questions in Table \ref{tab06}. Practically speaking, this systematic review serves as a guide for various researchers, practitioners, first responders, and crime analysts, to take into account the studied features and techniques to effectively understand and detect the crimes that, in part, foster the effort for early crime detection. 

\section{Conclusion} \label{sec 07}

The complexity of crimes has increased along with technological development, creating difficult problems for law enforcement. Researchers' interest in utilizing machine learning and deep learning to predict crime has increased recently, with an emphasis on finding patterns and trends in crime occurrences. In order to analyze the various machine learning and deep learning algorithms used in predicting crime, this paper looks at more than 150 articles. We have significantly studied the selected 51 articles to extract the essence of utilized various ML and DL techniques along with the publicly available datasets. 
The use of machine learning and deep learning algorithms to anticipate or identify criminal activity has shown significant promise in resolving the crime detection problem. These advances may help to increase the precision and efficacy of crime prediction models by leveraging large datasets and sophisticated algorithms. Although there is a lack of literary wisdom on how these technologies can be used to solve the problem of crime prediction, despite the advancements in this sector. Thus our findings help to understand the implications of various ML and DL techniques. Also, our mentioned datasets and future directions will help the existing research community to pursue their research in the area of crime prediction.

\section*{Declarations}

\begin{itemize}
\item Conflict of interest

The authors of this article have no conflict of interest to declare.
\end{itemize}

\clearpage
\begin{appendices}

\section{Additional papers on crime detection using Machine Learning and Deep Learning}\label{secA1}
\begin{longtable}{L{0.4\textwidth}L{0.15\textwidth}
   L{0.15\textwidth}L{0.175\textwidth}}
    \caption{Crime detection papers with data science algorithms}\label{tab a1}%
    \\ \hline
        Method & Algorithms Applied & Data Set & Performance \\ \hline
        Applied six different types of machine learning algorithms using similar characteristics on crime data. \cite{tamilarasi2020diagnosis} & NB, CART, KNN, LoR and SVM &  2001 to 2012 with more than 2,00,000 records & Aggregate Accuracy - 30\% \\ \hline
        K-nearest neighbor algorithm is applied to predcit the crime. \cite{kumar2020crime} & KNN & Police website of a city Indore in Madhya Pradesh & Aggregate Accuracy - 99.51\% \\ \hline
        Statistical modeling approaches are applied to predict the crime rate of a particular district of India using the previous year's statistics. \cite{agarwal2018crime} & Weighted Moving Average, Functional Coefficient Regression and Arithmetic Geometric Progression & Crimes in India 2001 - 2013 & Between the range 85\% and 90\% \\ \hline
        Predicts the description of the perpetrator using algorithms like Multilinear Regression, K nearest neighbors, and Neural Network. \cite{shermila2018crime} & KNN & San Francisco Homicide dataset (1981-2014) & 91.6\% \\ \hline
        Machine-Learning predictive models, K-nearest neighbor, and boosted decision trees are implemented to predict crime. \cite{kim2018crime} & KNN and boosted decision tree & Vancouver crime data for the last 15 years & Accuracy between 39\% to 44\% \\ \hline
        Spatial analysis and auto-regressive models to detect high-risk crime regions in urban areas and reliably forecast crime trends in each region. \cite{catlett2018data} & Association rule mining , classification, and clustering & Chicago & Mean Absolute Error (MAE) - 11.5 \\ \hline
        An auto-regressed temporal and a feature-based inter-area spatial correlation are applied to measure such patterns for crime prediction. \cite{yi2018integrated} & Gaussian Process Regressor, RF, Auto Regression Moving Average, Linear Regression and  Clustered Continuous Conditional Random Field  & Chicago from Jan. 1 2013 to Jan. 1 2016. & Aggregate RMSE - 3.1 \\ \hline
        Experimented with polynomial, auto-regressive, and support vector regression methods and found that the quality of support vector regression significantly outperforms other approaches.\cite{dash2018spatio} & Polynomial, auto-regressive and support vector regression & Different regions in City of Chicago & RMSE-  0-10 for most crimes \\ \hline
        Representative covariates from the nonhistorical crime data are added to the prediction model to explore the changes in the result accuracy of crime prediction. \cite{yao2020prediction} & Random Forest & Crime data set in San Francisco & Demo, POI, DI - 1.18 \\ \hline
        Spatio-Temporal Crime Hot-Spots detection based on clustering method. \cite{baqir2020evaluating} & Hierarchical Density-based spatial clustering of applications with noise (HDBSCAN) & NYC & Approx 10 sec for cluster \\ \hline
        Check crime rate based on text mining on social media using a logistic regression algorithm. \cite{permana2021crime} & Logistic Regression & Twitter \& Facebook & Aggregate Accuracy - 90\% \\ \hline
        Latitude and Longitude are used by KNN. In addition, Date is used by the naive bayes algorithm for classifying crime type. \cite{toppireddy2018crime}  & KNN \& Naive Bayes (NB)  & UK crime data (2015 to 2017).   & Accuracy - 70 to 80\%  \\\hline
        A Text-mining approach for crime tweets in Saudi Arabia. \cite{algefes2022text} & CNN, KNN, NB, SGD, SVM, DT and RF & Tweets & Aggregate Accuracy - 79\% \\ \hline
        Presents an empirical analysis of machine learning algorithms for crime prediction using stacked generalization, which is an ensemble approach. \cite{sivanagaleela2019crime} & DT, RF, SVM and KNN & 2001-2015 Crime Dataset in India & Accuracy - 99.5\% \\ \hline
        Proposes using auto-regression techniques for predicting crime in a city using time series data. \cite{yadav2018crime} & ARIMA and Generalized Linear Model & Crime Data set from India & Not Quantified \\ \hline
         Proposes a Spatial-Temporal Self-Supervised Hypergraph Learning framework (ST-HSL) to tackle the label scarcity issue in crime prediction. \cite{li2022spatial} & ST-HSLARIMA, SVM, STResNet, DCRNN, STGCN and GWN & NYC and Chicago Crime data & MAE 0.79 for robbery \\ \hline
        Proposes an artificial neural network (ANN) based approach for detecting crime-related posts on Twitter. \cite{sandagiri2020detecting} & Neural Network and SVM & Tweets & Accuracy - 90.33\% \\ \hline
        Machine learning algorithms to predict crimes that endanger public health and aim to improve the efficiency of crime prevention. \cite{wang2022preventing} & SVM and Random Forest & Records of crimes against public health in a chinese city from January 1, 2018 to October 31, 2018.  & Not Quantified \\ \hline
        Develop scalable predictive models for violent offending following (12 to 24 months post-discharge) discharge from secure psychiatric hospitals. \cite{wolf2018prediction} & Cox Regression & 2248 individuals discharged from Swedish psychiatric hospitals between 1992 and 2013 & Aggregate Concordance Index 0.73 \\ \hline
        Application of two different machine learning classifier models on a rich collection of real-world forensic casework images of drug-related offenses. \cite{abraham2021automatically} & SVM with Bag of Visual words and Tree-CNN & Australian Federal Police illicit drug database. 97287 images & True positive rate - 89.17\% \\ \hline
        Investigates the HDBSCAN to detect the hot spots that have a higher risk of crime occurrence and Seasonal Auto-Regressive Integrated Moving Average (SARIMA) is exploited in each dense crime region to predict the number of crime incidents in the future with spatial and temporal information. \cite{butt2021spatio} & HDBSCAN and SARIMA & NYC crimes 2008 to 2017. & Aggregate MAE - 11.7 \\ \hline
        
   \end{longtable}




\end{appendices}

\clearpage
\bibliography{Crime_ref}
\bibliographystyle{sn-standardnature}

\end{document}